\documentclass[11pt,a4paper]{article}
\usepackage{sumvae}
\usepackage{times}
\usepackage{latexsym}
\usepackage{amsfonts}
\usepackage{amsmath}
\usepackage{pgfplots}
\usepackage{url}

\aclfinalcopy 

\title{Unsupervised Abstractive Sentence Summarization using Length Controlled Variational Autoencoder}

\author{Raphael Schumann \\
  Institute for Computational Linguistics \\
  Heidelberg University \\
  {\tt rschuman@cl.uni-heidelberg.de}}

\date{}

\begin{document}
\maketitle
\begin{abstract}
In this work we present an unsupervised approach to summarize sentences in abstractive way using Variational Autoencoder. VAE are known to learn a semantically rich latent variable, representing high dimensional input. VAEs are trained by learning to reconstruct the input from the probabilistic latent variable. Explicitly providing the information about output length during training influences the VAE to not encode this information and thus can be manipulated during inference. Instructing the decoder to produce a shorter output sequence leads to expressing the input sentence with fewer words. We show on different summarization data sets, that these shorter sentences can not beat a simple baseline but yield higher ROUGE scores than trying to reconstruct the whole sentence. 
\end{abstract}

\section{Introduction}
The increasing amount of text data in the digital age calls for methods to reduce reading time while maintaining information content. The process of summarization achieves this by deleting, generalizing or paraphrasing fragments of the input text. Summarization methods can be categorized into single or multi document and extractive or abstractive approaches. In contrast to single document~\cite{rush}, the multi document setup can leverage the fact that in some domains like news articles there are different sources describing the same event~\cite{banerjee, haghighi}. Extractive methods solely rely on the words of the input and e.g. extract whole sentences~\cite{erkan, parveen} or recombine phrases on the sentences level~\cite{banerjee}. Abstractive approaches on the other hand are rarely bound to any constraints and gained a lot of traction due to recent advances in machine translation like the encoder-decoder framework~\cite{encdec, paulus} or attention mechanism~\cite{attn, rush, paulus}. Another more general distinction is the need of supervision. Supervised methods require training pairs of input text and output summarization~\cite{paulus, rush}, whereas unsupervised methods abuse inherent properties of the input like frequency of phrases~\cite{banerjee} or centrality~\cite{erkan}. In this work we use a Variational Autoencoder (VAE)~\cite{kingma, bowman} and control the decoding length~\cite{kikuchi} to obtain a shortened version of an input sentence. VAEs work unsupervised and decoding makes use of the whole available vocabulary. This work is organized into following sections. At first we give background about used technologies and concepts. In~\ref{model} we describe the architecture of our model. The data we use for the experiments in section~\ref{exp} is outlined in section~\ref{data}. At last we report the results in section~\ref{res}.

\section{Background}
\subsection{Variational Autoencoder}
Variational Autoencoder (VAE) is a generative model firstly introduces by~\cite{kingma}. Like regular autoencoders VAEs learn a mapping $q(z|x)$ from high dimensional input $x$ to a low dimensional latent variable $z$. Instead of doing this in a deterministic way VAE imposes a prior distribution on $z$, e.g. standard Gaussian:

\begin{equation}
    p(z)=\mathcal{N}(z; 0, 1).
    \label{pz}
\end{equation}

The desired effect is that each area in the $z$ space gets a semantic meaning and thus samples from $p(z)$ can be decoded in a meaningful way. The decoder $p_\theta(x|z)$ is trained to reconstruct the input $x$ based on the latent variable $z$. In order to approximate $\theta$ via gradient descent the \textbf{reparameterization trick}~\cite{kingma} was introduced. This trick allows the gradient to flow through the sampling decision of $z$ (Formula~\ref{pz}) by outsourcing the discrete operation. Let $\mu$ and $\sigma$ be deterministic outputs of the encoder $q_\theta(\mu, \sigma|x)$:
\begin{equation}
    z = \mu + \sigma \odot \epsilon \quad \textnormal{where } \epsilon \sim \mathcal{N}(0, 1)
\end{equation}
and $\odot$ is the element-wise product. To prevent the model pushing $\sigma$ close to $0$ and basically fall back to a regular autoencoder the objective is extended by the Kullback-Leibler (KL) divergence between prior $p(z)$ and $q(z|x)$:

\begin{equation}
\begin{split}
    \mathcal{L}(\theta; x)=-KL(q_\theta(z|x)||p(z))\\
    + \mathbb{E}_{q_\theta(z|x)}[log p_\theta(x|z)].
\end{split}
\label{obj}
\end{equation}

The goal is to have a non-zero, but not out of control KL term while maintaining a reasonable reconstruction loss. This guarantees a semantically rich latent variable and good generation ability.

\subsection{Controlling Output Length} \label{lemb}
There are different methods for controlling the output length in an encoder-decoder model. One of them is \textit{LenEmb}~\cite{kikuchi} where the decoder is fed information about the remaining length at every decoding step $t$. This information is encoded as an embedding matrix $W_{L}$ accessed by $L_{l_t}= e_L(l_t)$ and learned during training. Instead of calculating the remaining length as bytes we use a more straight forward approach by counting whole words. At each decoding step the length embedding is concatenated to the input and chosen as follows:

\begin{equation}
    \begin{array}{ll}
         l_1 = length \\
        l_{t+1} = \max \{l_t-1, 0\},
    \end{array}
\end{equation}

where $length$ is the desired length. This encourages the decoder to fit the information left into the remaining words. The authors show in a supervised summarization setup that setting $length$ to the desired number of output bytes, conveniently the 75 bytes of the references, yield better performance during evaluation.

\section{Model} \label{model}
\begin{figure}
    \centering
    \includegraphics[width=0.45\textwidth]{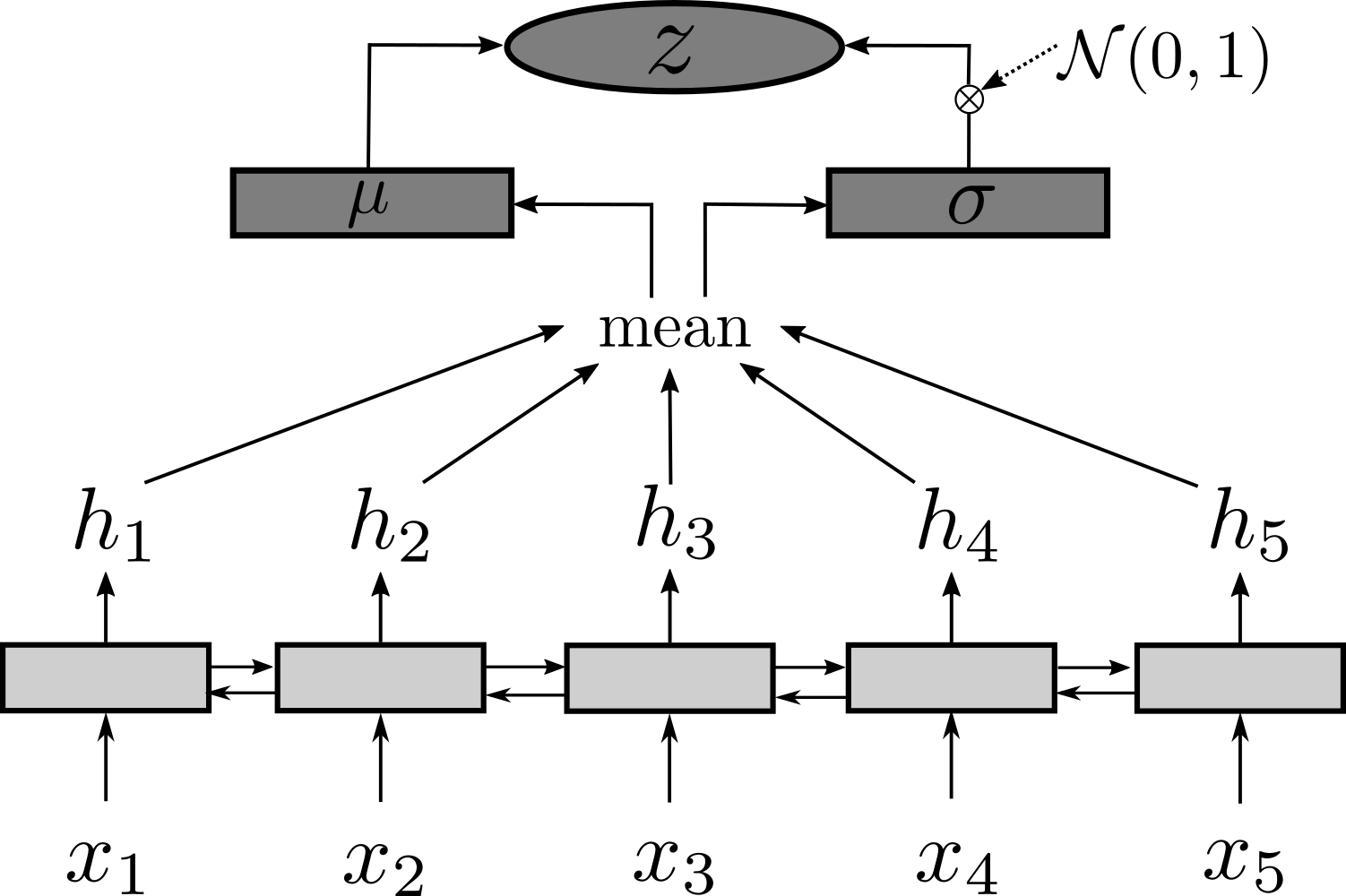}
    \caption{VAE Encoder with bidirectional RNN and mean representation of the input}
    \label{encoder}
\end{figure}

\begin{figure}
    \centering
    \includegraphics[width=0.45\textwidth]{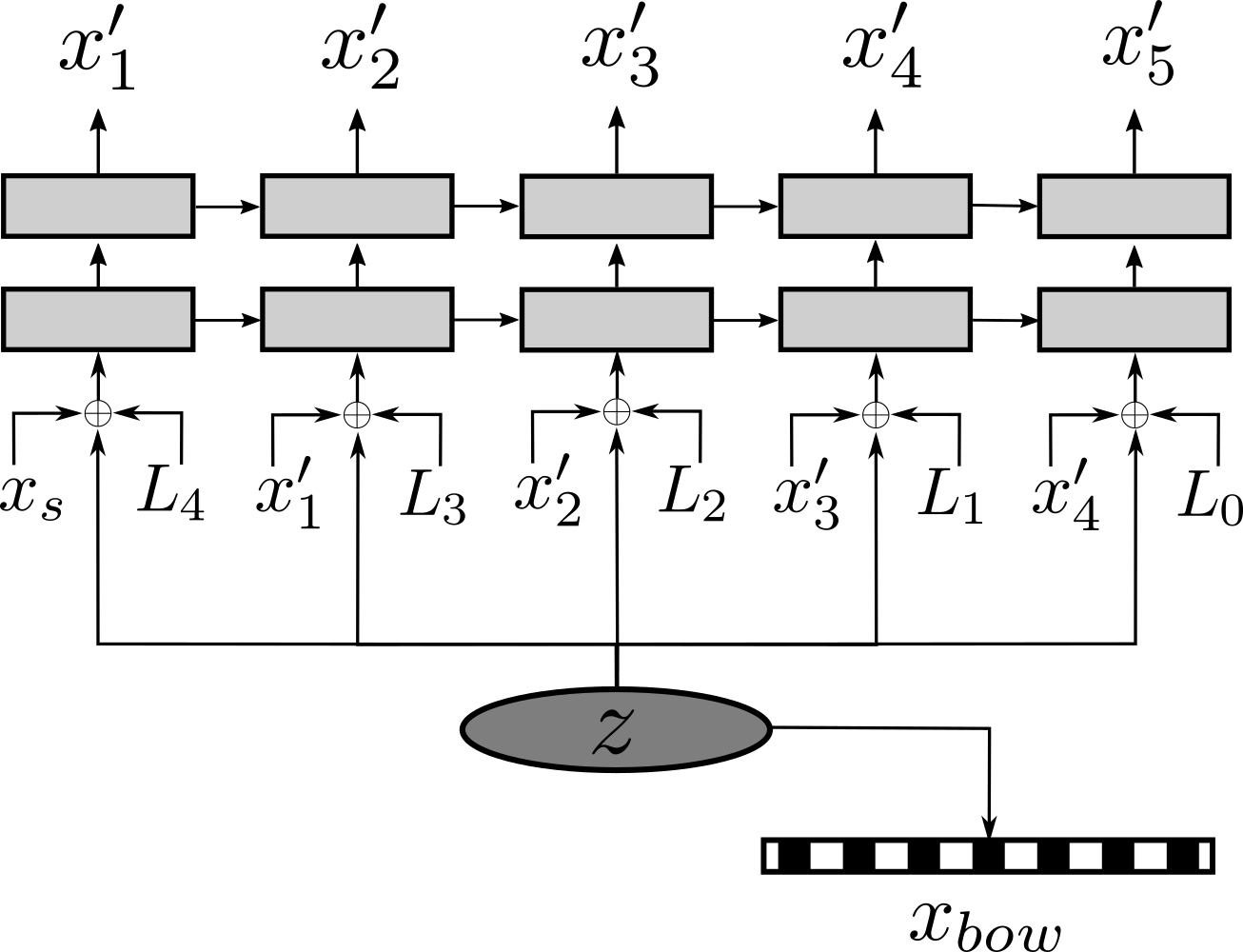}
    \caption{VAE Decoder with \textit{bag-of-word loss} and \textit{LenEmb}}
    \label{decoder}
\end{figure}

In order to apply the VAE principle to text data,~\cite{bowman} employ RNNs as encoder and decoder. The vectors $\mu$ and $\sigma$ are constructed from the last hidden state of the encoder and the first cell state of the decoder is initialized as $z$. Since then many improvements of this basic architecture have been published and are adopted in this work. First of all we use a bidirectional encoder which reads forward and backward through the input sequence $x$. At each encoding step the forward and backward hidden states $fh_i$ and $bh_i$ are concatenated to $h_i=[fh_i, bh_i]$. ~\cite{vani} then calculate $\mu$ and $\sigma$ from the mean of all hidden states $h_i$, arguing that this produces a better sequence representation and the gradient reaches every input vector more easily. This is depicted in Figure \ref{encoder}. Besides the reconstruction loss of the input sequence ~\cite{zhao} introduce a so called \textit{bag-of-words loss}. A $V$ dimensional vector is predicted by a feed-forward layer which takes $z$ as input, where $V$ is the vocabulary size. This vector is compared against the label $x_{bow}$ which is the one-hot representation of the input sentence. This forces the model to put more general information into the latent variable instead of encoding the start of a sentence and derive the rest by memorizing word order in the decoder. As seen in Figure~\ref{decoder} the multi-layer RNN gets fed the latent variable at every decoding step, again allowing to have an easier way for the gradient to flow back. Additionally the last emitted word $x'_{t-1}$ and the length embedding, see~\ref{lemb}, are concatenated to the input. To speed up the training sampled softmax~\cite{sampledsoft} estimates the softmax function at each decoding output.


\section{Data} \label{data}
The data setup is similar to ~\cite{rush}. For training they use 4 million pairs of title and first sentence of the article from Gigaword~\cite{gigaword} data set. As we do not need supervision we remove the titles and due to resource limitations remove all sentences with more than 30 words. The remaining 1.8 million training sentences are preprocessed by lower-casing and tokenizing all words. Additionally numbers are replaced by \# and words not in the top 40000 are replaced by UNK token. For evaluation we also use the around 2000 held-out article-title pairs from Gigaword and the DUC-2004 set~\cite{duc2004}. This consist of 500 news articles from New York Times and Associated Press Wire service and comes with 4 different reference summaries (capped at 75 bytes) written by humans. 

\section{Experiment} \label{exp}

We train the proposed model on the above presented data by maximizing the objective in Formula~\ref{obj}.
To obtain a shortened version of the input sentence during testing we set $l_1$ to the desired length. Our assumption is that the decoder tries to fit all the information present in the latent variable into the limited output words. Doing so by skipping meaningless words or rephrasing semantic bits to fewer tokens. All under observation of the implicit language model ensuring a grammatically correct sentence.

\subsection{Baseline}
We use \textsc{Prefix} as baseline which cuts the first 75 characters from the input sentence as summarization. This simple baseline shows to what extent out model is able to pass the information of the input sentence trough the low dimensional latent variable.
\subsection{Training Details}
Similar to ~\cite{bowman} a weight for the KL term in the objective function is annealed from 0 to 1 during training. This hinders the model to go the easy way and set the KL term to $0$ by letting $q_\theta(z|x)$ be equal to $p(z)$. This would mean there is no information encoded in $z$ and degenerate the VAE to a regular language model. Another technique to overcome this is dropping the previous emitted word during decoding, relying the decoder further on the latent variable.\par
The LSTM cell~\cite{lstm} is used as basic RNN unit. Optimization is done by Adam~\cite{adam} and sampled softmax draws 1000 words. Beam search size is set to 100 and batch size to 512. The number of desired output words is set to 20 to reliable reach the 75 bytes of the reference summarizations. All other hyperparameter are searched by Bayesian optimization\footnote{\url{https://scikit-optimize.github.io/}}. Encoder and decoder RNN cell size is 243. Word embedding size is 254 and the latent variable has 124 dimensions. A 236 wide hidden layer predicts $X_{bow}$. The best size for length embeddings is found to be 50. Words are not dropped during decoding by a probability of 0.20 and the output layer of RNN cells is regularized by a dropout keep rate of 0.87. 

\begin{figure}
    \centering
    \includegraphics[width=0.45\textwidth]{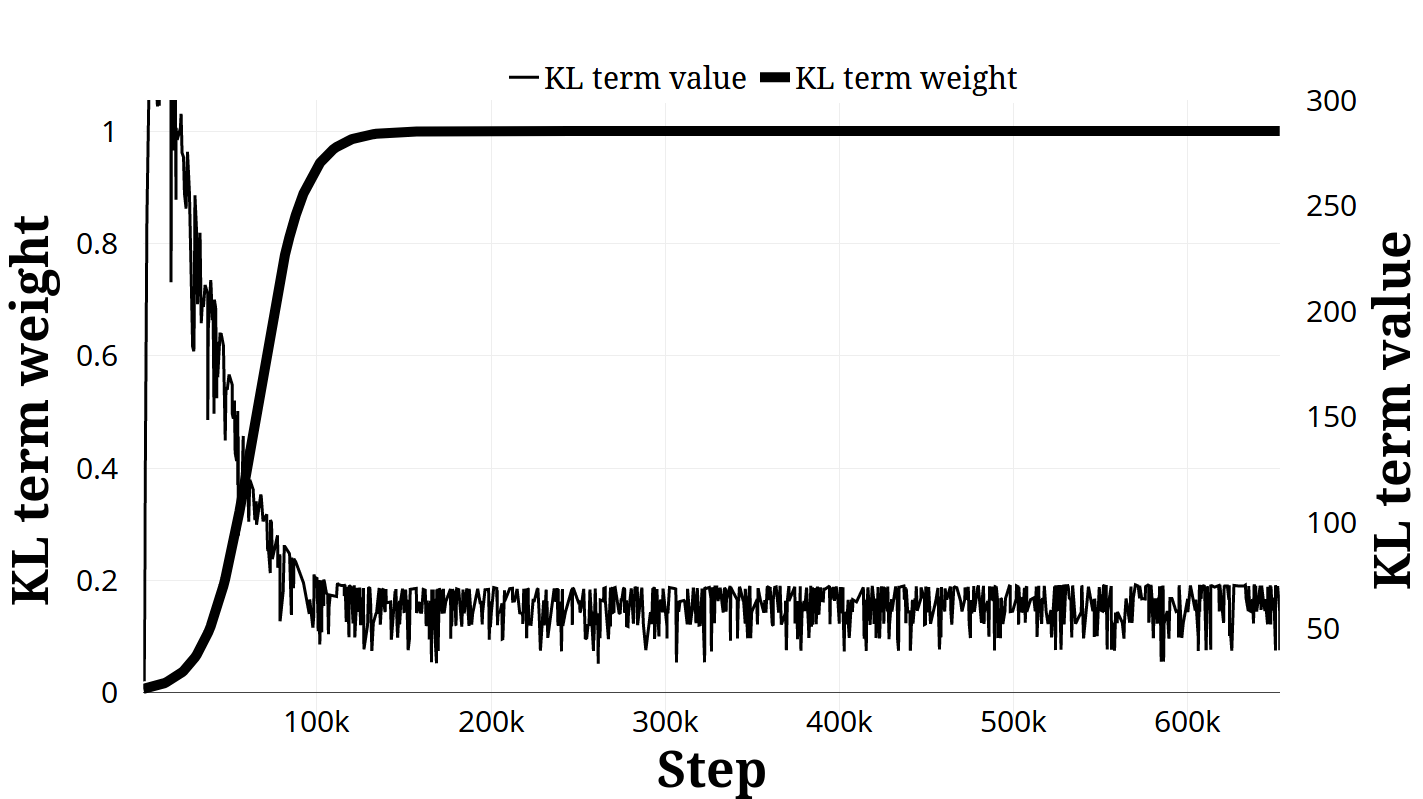}
    \caption{The annealing of the KL term weight during training steps and the reaction of KL term value}
    \label{klplot}
\end{figure}
\section{Results} \label{res}
\begin{table*}[ht]
\centering
\begin{tabular}{l|lll|llll}
\hline
\multicolumn{1}{c|}{} & \multicolumn{3}{c|}{DUC-2004} & \multicolumn{4}{c}{Gigaword}         \\
Model                 & ROUGE-1  & ROUGE-2  & ROUGE-L & ROUGE-1 & ROUGE-2 & ROUGE-L & Ext. \% \\ \hline
\textsc{Prefix}     & 22.43    & 6.49     & 19.65   & 23.14   & 8.25    & 21.73   & 100     \\
no len limit          & 14.49    & 2.06     & 12.28   & 19.91   & 4.14    & 18.02   & 51      \\
LenEmb 20             & 16.38    & 2.56     & 14.19   & 22.19   & 4.56    & 19.88   & 60     
\end{tabular}
\caption{ROUGE-1, ROUGE-2, ROUGE-L on DUC-2004 and Gigaword evaluation set. no len Limit decodes the input sentence with modifying the length. LenEmb 20 sets the desired length to 20 output words. Ext. \% reports the amount of extracted words from input.}
\label{res}
\end{table*}
\subsection{Evaluation Metric}
ROUGE~\cite{rouge} is an n-gram based evaluation metric to quantify the quality of a summary relative to given references. We report results on ROUGE-1 and ROUGE-2 which basically count the uni- and bi-gram overlap. Furthermore ROUGE-L score is based on the longest common subsequence (LCS) between the given texts. ROUGE is just an indicator if a automatically generated summary is as good as a human-written reference and should be handled with caution.
\subsection{Quantitative Evaluation}
\begin{figure}[h]
    \centering
    \includegraphics[width=0.45\textwidth]{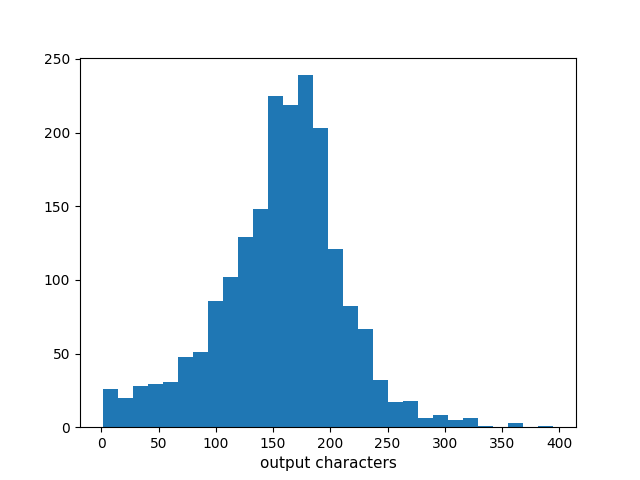}
    \caption{Frequency of output characters without limiting the length}
    \label{giganone}
\end{figure}

\begin{figure}[h]
    \centering
    \includegraphics[width=0.45\textwidth]{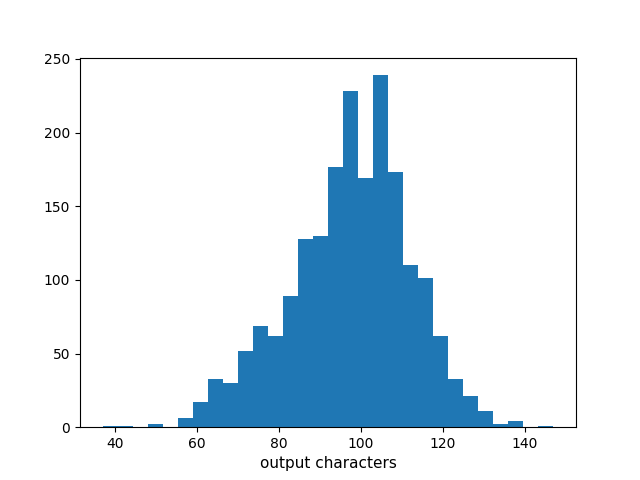}
    \caption{Frequency of output characters with setting desired length to 20}
    \label{giga20}
\end{figure}

Before discussing the summarization results we take a look at how the \textit{LenEmb} effects the model. In Figure~\ref{giganone} and~\ref{giga20} we see the output length of the model without length restrictions and the one with a desired length of 20 words. Figure~\ref{giganone} is about the same distribution as the input sentences. Figure~\ref{giga20} proofs that we are able to reduce the output length near the desired 75 characters. In fact 20 words are chosen to have the majority slightly above 75 characters to not waste word space during ROUGE evaluation. We perform another analysis to study the effect of \textit{LenEmb}. We train a model with explicitly providing the information about the sentence length via \textit{LenEmb} and one without this extension. This means the model has to somehow encode the length information into the latent variable to reproduce the input sentence with minimal loss. In Table~\ref{r2} we see the $R^2$ results of a Linear Regression (LR) trained on the latent variables of both models with the objective to predict the length of the encoded sentence. For the model without explicit length information LR can better predict the length of the encoded sentence with only looking at the latent variable. With less length information stored in the latent variable it should be easier to influence the model to produce a certain output length.\\
The ROUGE scores are found in Table~\ref{res}. Our model is not able to beat the \textsc{Prefix} baseline. This however could be the effect of the VAE not being able to restore the correct input sentence. We verify this by testing a vanilla VAE model on solely reconstructing the input sentence and see that a lot of mistakes are made. One reason is the lack of attention, which can't be used in a VAE setting, to 'copy' rare words from the input. Our \textit{LenEmb} model however is consistently better than the vanilla VAE, which shows that the reducing of output length can fit more information into the first 75 characters. If we could improve the vanilla VAE to reproduce the input sentence without making a lot mistakes and the \textit{LenEmb} model maintains the performance gain over the vanilla VAE, we could beat the \textsc{Prefix} baseline. The grammatical quality of the generated sentences was not evaluated.


\begin{table}[h]
\centering
\label{len}
\begin{tabular}{lll}
$R^2$                                       & DUC2004                   & Gigaword \\ \hline
\multicolumn{1}{l|}{with \textit{LenEmb}} & \multicolumn{1}{l|}{0.41} & 0.54     \\
\multicolumn{1}{l|}{w/o \textit{LenEmb}}  & \multicolumn{1}{l|}{0.59} & 0.72    
\end{tabular}
\caption{Linear Regression prediction on sentences length with and w/o \textit{LenEmb}}
\label{r2}
\end{table}

\section{Conclusion}
We extended a VAE with \textit{LenEmb} to control the length of the produced sentences. The hypotheses that stimulating the decoder to produce shorter outputs will result in more information expressed in fewer words could be verified in a summarization experiment. However a simple baseline could not be beaten with this approach. A reason and subject to further research is how the vanilla VAE can be improved to better reconstruct the input sentence and how this influences the \textit{LenEmb} extended model. A Linear Regression experiment demonstrated that the length of the input sentence is encoded in the latent variable. All in all this is a reasonable approach to construct a unsupervised abstractive sentence summarization model and worth further investigation.

\bibliography{sumvae}
\appendix

\end{document}